\documentclass[a4paper,twoside]{article}
\usepackage{epsfig}
\usepackage{subcaption}
\usepackage{calc}
\usepackage{amssymb}
\usepackage{amstext}
\usepackage{amsmath}
\usepackage{amsthm}
\usepackage{multicol}
\usepackage{pslatex}
\usepackage{apalike}
\usepackage{color}
\usepackage{url}
\usepackage{graphicx}
\usepackage{paralist}

\usepackage{censor}
\usepackage[latin1]{inputenc}

\usepackage{SCITEPRESS}     
\newcommand{\bv}{\textit{\textbf{v}}}

\usepackage{xcolor}
\newcommand{\rev}[1]{{\color{black} #1}}

\begin{document}
\StopCensoring
\title{On the Choice of General Purpose  Classifiers in  Learned Bloom Filters: An Initial Analysis Within Basic Filters}

\author{
  \authorname{
    \censor{Giacomo Fumagalli}
    \sup{1}
    \orcidAuthor{0000-0002-2068-9293},
    \censor{Davide Raimondi}
    \sup{1}
    \orcidAuthor{0000-0001-8171-8302},
    \censor{Raffaele Giancarlo}
    \sup{2}
    \orcidAuthor{0000-0002-6286-8871},
    \censor{Dario Malchiodi}
    \sup{1,3}
    \orcidAuthor{0000-0002-7574-697X}
    and \censor{Marco Frasca}
    \sup{1,3}
    \orcidAuthor{0000-0002-4170-0922}
  }
\affiliation{\sup{1}Dipartimento di Informatica, Universit\`{a} degli Studi di Milano, 20133 Milano, Italy}
\affiliation{\sup{2}Dipartimento di Matematica ed Informatica, Universit\`{a} di Palermo, 90133 Palermo, Italy}

\affiliation{\sup{3}CINI National Laboratory
in Artificial Intelligence and Intelligent Systems (AIIS), Universit\`{a} di Roma, 00185 Roma, Italy}

\email{\xblackout{\{giacomo.fumagalli1,davide.raimondi2\}@studenti.unimi.it, raffaele.giancarlo@unipa.it, dario.malchiodi@unimi.it, marco.frasca@unimi.it (corresponding author)}}
}


\abstract{Bloom Filters are a fundamental and pervasive data structure. Within the growing area of Learned Data Structures, several Learned versions of Bloom Filters have been considered, yielding advantages over classic Filters. Each of them uses a classifier, which is the Learned part  of the data structure. Although it  has a central role in those new filters, and its space footprint as well as classification time may affect the performance of the Learned Filter,   no systematic study of which specific classifier to use in which circumstances is available. We report progress in this area here, providing also initial guidelines on which classifier to choose among five classic classification paradigms.}

\onecolumn \maketitle \normalsize \setcounter{footnote}{0} \vfill

\section{\uppercase{Introduction}}
\label{sec:introduction}

Learned Data Structures is a  novel area at the crossroad of Classic Data Structures and Machine Learning. They have been initially proposed by Kraska et. al. \cite{Kraska18} and have had a rapid growth
\cite{Ferragina:2020book}. Moreover, now the area  has been extended to include also Learned  Algorithms \cite{Mitz20}.     
The theme common to those new approaches to Data Structures Design and Engineering is that a query to a data structure is either intermixed with or preceded by a query to a Classifier \cite{duda20} or a Regression Model~\cite{FreedmanStat}, those two being the learned part of the data structure.  To date, 
	Learned Indexes have been the most studied, e.g.,  \cite{amato2021learned,Ferragina:2020pgm,FERRAGINA21,Kipf20,Kraska18,Mailtry21,Marcus20,Markus20b}. Rank/select data structures have also received some attention \cite{Boffa21}.

	Bloom Filters \cite{Bloom70}, which are the object of this study, have also been considered,  as we detail next. Historically, they have been designed to be a data structure able to solve the Approximate Membership  Problem in small space (see Section \ref{sec:ASM}). Due to their fundamental nature and pervasive use, many variants and alternatives have been proposed. A good review in the domain of Internet applications is provided in \cite{Broder2005}.

	Kraska et al. \cite{Kraska18} have proposed a learned version of such a filter, in which a query to the data structure is preceded by a query to a suitably trained  binary classifier. The intent is to reduce space and reject time. Mitzenmacher \cite{Mitz18}  has provided a model for those filters, together with a very informative mathematical analysis of their pros/cons and even an alternative proposal. Additional Learned Bloom Filters have been proposed  recently \cite{Dai,Kraskap}. It is worth pointing out that, although they differ in architecture, each of those proposals has a classifier as its central part. 
	
	Somewhat puzzling is that, although the classifier is the novel part of this new family of Filters and it is accounted for in theoretical studies \cite{Mitz18}, not much attention is given to which classifier to use in practical settings. Kraska et al. use  a Neural Network, while Dai and Shrivastava and Vaidya et al. use Random  Forests. Those choices are only informally motivated, giving no evidence of superiority with respect to other possible ones, via a comparative analysis. 
	Therefore, the important problem of how to choose  a specific classifier in conjunction with  a specific filter has not been addressed so far,  even in one application domain.

     Such a State of the Art is problematic, both methodologically and practically. Here we propose the first, although initial, study that considers the choice of the classifier within Learned Filters. Since our study is of a fundamental nature, we consider only basic versions of the Filters and six generic, if not textbook, classifier paradigms.
     The intent is to characterize how different classifiers affect different performance parameters of the Learned Filter.  The application domain is the one of malicious URL determination, as in previous work. \rev{As per literature standard, our datasets are real and fall within the  data size used by Dai and Shrivastava}. However, even so, Random Forest classifiers do not seem to be competitive in our setting, needing bigger datasets. Indeed, for the continuation of this work, we plan to consider much larger datasets and include also Random Forests. Additional application domains will also be considered.  
     
     Our results provide a confirmation of some expected behaviours, like that more complex classifiers tend to yield larger space reduction when used within a learned Bloom Filter, but also some counterintuitive perspectives, including that simple linear classifiers might represent very competitive alternatives to their bigger counterparts, sometimes even better. These results make more sense if we consider that  in a learned Bloom Filter classifiers come up with one or two classic Bloom Filters, calibrated in turn on the classifier performance, and accordingly it is their synergy that determines the overall space reduction, not just the classifier itself. Finally, a relevant note is also that in our experiments linear classifiers allow two orders of magnitude faster reject times.




\section{Approximate Set Membership Problem: Definitions and Key Parameters}\label{sec:ASM}
\paragraph{{\bf Definition}.} Given $S\subset U$ and $x\in U$, where $U$ is the universe of keys, the \textit{set membership} problem consists in finding a data structure able to determine if $x \in S$. The \textit{approximate
set membership} instead allows for a fraction $\epsilon$ of \textit{false positives}, that is elements in $U \setminus S$ considered as members of $S$. No \textit{false negatives} are allowed.  That is,  elements in $S$ considered as non-members.\\ 

\paragraph{\textbf{A  Paradigmatic Example}. } Assume that a set of URLs is given. They can be \emph{malicious}, i.e.,  websites hosting
unsolicited content (spam, phishing, drive-by downloads, etc.), or luring unsuspecting users to become victims of scams (monetary loss, theft of private information, and malware installation), or otherwise labeled as \emph{benign}. The aim is to design a time efficient data structure that takes small space and that ``rejects'' benign URL quickly, as they do not belong to the malicious set. On occasions, we may have false positives. 

\paragraph{\bf Key Parameters of the Filter.} The reject time, taken as the expected time to reject a  non-member of the set; the space taken by the data structure, since the cardinality of $S$ may be very large; the false positive rate $\epsilon$. Quite remarkably, Bloom provided two data structures to solve the posed problem, linking the three key parameters in trade-off bounds. The most space-efficient of those data structures goes under the name of Bloom Filter and it is outlined next, together with two of its learned versions. 

\subsection{Basic Classic and Learned Bloom Filters}\label{sub:BFs}

\paragraph{\bf Classic.}
Letting $|S|=n$, a Bloom Filter (BF) is made up by a bit array $\bv$ of size $m$, whose elements are all initialized to $0$. It uses $k$  hash functions $h_j: U \longrightarrow \{0, 1, \ldots, m-1\}$,  $j \in \{1, \ldots, k\}$, which can be assumed to be perfect in theoretic studies (see \cite{Mitz18}). 
When a new element $x$ of $S$ is added, it is coded using the hash functions: each bit in $\bv$ in position $h_j(x)$ is set to $1$, for each $j \in \{1, \ldots, k\}$. 
To test if a key $x\in U$ is a member of $S$, $h_j(x)$ for each $j$ is computed, and $x$ will be rejected if there exists $j \in \{1, \ldots, k\}$ such that $v[h_j(x)] = 0$. However, when the Bloom Filter considers $x$ as a member, it might be a false positive due to the hash collisions.  The false positive rate $\epsilon$ is inversely related to the space usage of the bit array. More accurately, the trade-off formula   connecting the key parameters of the Filter is given in equation (21) in \cite{Bloom70}. Analogous trade-off formulas are also known, e.g., \cite {Broder2005,Mitz18}. In  applications~\cite{Broder2005}, one usually asks for the the most space-conscious filter, given a false positive rate, being the reject time a consequence of those choices.


\medskip 

\paragraph{\bf Learned: One classifier and one filter.}
The intent of this data structure, named Learned Bloom Filter (LBF), is to achieve a given false positive rate, as in a classic filter, but in less space, and possibly with little loss in reject time~\cite{Kraska18}. One can proceed as follows. A classifier $C:U \longrightarrow [0,1]$ is trained on a labeled dataset 
$(x, y_x)$, where $x\in D \subset U$, $S\subset D$, and $y_x=1$ when $x \in S$,  $y_x=-1$ otherwise. The positive class is thereby the class of keys $S$. The larger $C(x)$, the more likely $x$ belongs to $S$. Then, in order to have a binary prediction, a threshold $\tau \in (0,1)$ is to be fixed, yielding positive predictions for any $x\in U$ such that $C(x) \geq \tau$, negative otherwise. 
To avoid false negatives in the learned Bloom Filter, that is elements $x\in S$ rejected by the filter, a classic ``backup'' Bloom Filter $F$ is created on the set $\{x \in S|C(x) <\tau\}$. A generic key $x$ is tested against membership in $S$ by computing the corresponding prediction $C(x)$: $x$ is considered as an element of $S$ if $C(x) \geq \tau$ or $C(x) < \tau$ and $F$ does not reject $x$. $x$ is rejected otherwise. Unlike a classic Bloom Filter, the false positive rate of a LBF depends on the distribution of a given query set $\bar S\in U\setminus S$~\cite{Mitz18}. Hereafter, we will refer to $\epsilon_{\tau} = |\{x \in \bar S |  (C(x)\geq \tau)\}|$ as the \textit{empirical} false positive rate of the classifier on $\bar S$, to $\epsilon_{F}$ as the false positive rate of $F$, and to the \textit{empirical} false positive rate of the LBF on $\bar S$ as $~\epsilon = \epsilon_{\tau} + (1-\epsilon_{\tau})\epsilon_{F}$. Fixed a desired $\epsilon$, the optimal value of $\epsilon_F$ is \[
\epsilon_F = \frac{\epsilon - \epsilon_{\tau}}{1 - \epsilon_{\tau}},\ 
\] which yields the constraint $\epsilon_{\tau} < \epsilon$. 

It is to be pointed out that now, false positive rate $\epsilon$, space and reject time of the entire filter are intimately connected and influenced by the choice of $\tau$. Another delicate point that emerges from the analysis  offered in \cite{Mitz18} is that, while the false positive rate of a classic Filter can be reliably estimated experimentally because  of its data independence, for  the Learned version this is no longer so immediate and further insights are needed. As  an additional contribution, an experimental methodology is  suggested in \cite{Mitz18}  and we adhere  to it here.

\medskip

\paragraph{\bf Learned: Sandwiched classifier.}
The intent of this variant, named Sandwiched Learned Bloom Filter (SLBF), is to filter out most non-keys before they are supplied to the LBF. This would allow the construction of a much smaller backup filter $F$ and an overall more compact data structure~\cite{Mitz18}. The specifics follow.  A Bloom Filter $I$ for the set $S$ precedes the structure of the Learned Bloom Filter described in the previous paragraph. The subsequent LBF is constructed on the elements of $S$ not rejected by $I$. Clearly, $I$ might yield a considerable number of false positives, when a limited budget of bits is dedicated to it. A query $x\in U$ is rejected by the SLBF if $I$ rejects it, otherwise the results of the subsequent LBF is returned. 
Denoted by $\epsilon_{I}$ the false positive rate of the filter $I$, the empirical false positive rate of a SLBF is $\epsilon = \epsilon_I \big(\epsilon_{\tau} + (1-\epsilon_{\tau})\epsilon_{F} \big)$. For a desired value of $\epsilon$, the following properties hold~\cite{Mitz18}:
\begin{itemize}
    \item $\epsilon_I=\frac{\epsilon}{\epsilon_{\tau}} \Big(1-\frac{FN}{n}\Big)$, with FN number of false negatives of $C$;
    \item $\epsilon\Big( 1-\frac{FN}{n}\Big) \leq \epsilon_{\tau} \leq 1-\frac{FN}{n}$.
\end{itemize}
As for the previous learned Bloom Filter variant, the classifier accuracy affects all the key factors of a SLBF, namely false positive rate, space and reject time. Considerations and experimental methodologies in this case are the same as in the previous case.

\section{Experimental Methodology}\label{sec:reprod}

\subsection{General Purpose Classifiers}\label{sub:classif}
The classifiers used in our analysis are \rev{briefly described below}. We assume each $x \in U$ is represented through a set of $n$ \rev{real-valued} attributes $A = \{A_1, \dots, A_n\}$, and that $Y = \{-1,1\}$ is the set of labels, with -1 denoting the negative class. The training set is made up by labeled instances $D\subset U$, in the form of couples $(x, y_x)$,  with $x\in D$ and $y_x \in Y$.  
\begin{itemize}
\item \textit{RNN-k}. A character-level recurrent neural network with Gated Recurrent Units (GRU)~\cite{RNN}, having $10$-dimensional embedding and $k$-dimensional GRU. RNN is included as baseline from literature for the same problem~\cite{Kraska18}. The parameters to be learned are the connections weights and unit biases for all the layers in the model.  

\textit{Hyperparameters}. The hyperparameters here are the embedding dimension, the GRU size $k$, the learning and the dropout rate. No changes in their configuration has been made with regard to the setting used in~\cite{Kraska18}, except for the size of the embedding layer, reduced to $10$, as done in \cite{ma2020}, to comply with the reduced size of our dataset, and the smaller size of the overall filter as well. The embedding dimension and $k$ affect the design of the Learned Filters. Therefore, we refer to them as  key hyperparameters. We use the same terminology also for the other classifiers used in this study. 
\\
\item \textit{Naive Bayes (NB)}. A Naive Bayes Classifier~\cite{NaiveBC} ranks instances based on Bayes' rule. It learns the conditional probabilities of having a certain label $Y=y_k$ given a specification of the attributes $A_i$:
\small
\begin{displaymath}
Y \leftarrow \arg\max\limits_{y{_k}} \frac{P(Y=y_k)P(A_1=x_1\ldots A_n=x_n|Y=y_k)}{\sum\limits_{j}P(Y=y_j)P(A_1=x_1\ldots A_n=x_n|Y=y_j)}. 
\end{displaymath}
\normalsize
The conditional probabilities on the right are parameters of the model, and are efficiently estimated from training data by assuming all attributes are conditionally independent given $Y$ (naive). To deal with missing attribute configurations and to avoid zero estimates of some parameters, a regularization yielding smooth estimates is applied. 

\textit{Hyperparameters}. The only hyperparameter is the one determining the strength of the smoothing for missing combinations of the attributes, which is not a key hyperparameter.
\item \textit{Logistic Regression (LR)}. The logistic regression~\cite{LRegr} classifies an instance $x = (x_1, \ldots, x_n)$ as negative if
\begin{displaymath}
P(Y=-1|A=x) > P(Y=1|A=x), 
\end{displaymath}
where
\begin{displaymath}
P(Y=-1|X=x) = \frac{\exp(w_0 + \sum_{i=1}^n w_ix_i)}{1 + \exp(w_0 + \sum_{i=1}^n w_ix_i)}, 
\end{displaymath}
and $P(Y=1|X=x) = 1 - P(Y=-1|X=x)$. Here, $w = w_0, \ldots w_n$ are parameters of the classifiers, which are estimated from training data. A regularization is typically applied to impose the norm of $w$ to be minimized along with the maximization of the conditional data likelihood. 

\textit{Key hyperparameters}. The model has just one hyperparameter, the one determining the strength of the regularization, which is not key. 
\item \textit{Linear Support Vector Machine (SVM)}. Linear SVMs~\cite{SVM} classify a given instance $x\in\mathbb{R}^n$ as $f(x) = w^T\cdot x +b$, where $\cdot$ is the inner product operator, $w \in \mathbb{R}^n$, and $b \in \mathbb{R}$. The hyperplane $(w, b)$ is learned to maximize the margin between positive and negative points: \begin{displaymath}\label{eq:SVM} 
\begin{array}{l}
	{\displaystyle \min_{\substack{w \in \mathbb{R}^n}} \frac{1}{2}\|w\|^2 +\ C\sum_{x\in D}\xi_x}
\\
	\text{s.t.} \quad y_xf(x) \geq 1-\xi_x  \quad x \in D
\\
\	\quad\hspace*{3mm} \xi_x \geq 0  \hspace{1.65cm} x \in D~,
\end{array}
\end{displaymath}
where $\xi_x$ is the error in classifying instance $x$, and $C$ is an hyperparameter regulating the tolerance to misclassifications. Non-linear SVMs, e.g. with Gaussian kernel, have been discarded due to their excessive size, in light of the fact that they need to store not just the hyperplane, but the also the kernel matrix.

\textit{Hyperparameters}. $C$ is the unique hyperparameter for this classifier, and it is not key. 
\item \textit{FFNN-h}. A Feed-Forward neural network~\cite{FFNN} with the input layer of size $n$, one hidden layer with $h$ units, and one output unit with sigmoid activation. The weights of connections and the neuron biases are the parameters to be estimated from training data.  

\textit{Hyperparameters}. The learning rate, the activation function for the hidden layer, and the number of hidden units $h$ are hyperparameters here, with the latter being also a key hyperparameter. 

\end{itemize}


\subsection{Data and Hardware}
Following the literature, we concentrated on URL data. URLs are  divided into  benign and malicious addresses, with the latter typically stored in the filter. They are as follows.

\begin{itemize}
    \item The first dataset comes from~\cite{URLUnimas}, and contains 15k benign and 15k malicious URLs.
\item The second dataset comes from~\cite{URLTrento}, in particular as benign we have chosen 3637 entries in the category legitimate uncompromised, and as malicious 38419 entries in the category hidden fraudolent.
\end{itemize} 
As for the training of the classifiers, we use both the mentioned datasets with their division  in benign and malicious. Since the number of benign elements does not guarantee good training on the NN-based  classifiers included in this research, we also consider the following dataset.  

\begin{itemize}
    \item All the 291753 URL available at~\cite{URLBotw}, that we take as being benign.
\end{itemize}

In analogy with~\cite{ma2020}, we processed the Web addresses as follows.
\begin{itemize}
    \item \textit{Standardization}. Common strings like {\tt "www."} and {\tt "http://"}  have been removed from each URL, and URLs have been resized to a length of 150, by either padding shorter addresses with a marker symbol or by truncating excess characters, removing duplicated URLs. The resulting dataset contains $310329$ benign and $43744$ malicious URLs.
\end{itemize}

Finally, the datasets have been suitably coded for the input of the classifiers employed in this work. 
\begin{itemize}
    \item \textit{Input coding for RNN}. In analogy with~\cite{ma2020}, URLs have been coded by mapping
each character to consecutive integers starting from 1 to 128 in order of frequency in the training set, and to 0 all the remaining characters, eventually obtaining a 150-dimensional vector. For their nature, such classifiers need an embedding layer taking as input the whole sequence of characters in the address to execute the training.
    \item  \textit{Input coding for other classifiers}. Here a standard \textit{bag of char} coding is adopted, based on the fact that the remaining classifiers are more flexible on their input format (no embedding needed). In this coding, each URL is represented with a vector containing for each distinct character in the training set its frequency in the URL--we tested both relative and absolute frequency, with no significant differences in the classifiers' performance. The extracted vectors are positional, in the sense each distinct character is assigned a fixed position in the vector. This way, addresses are coded with vector of dimension $79$. We point out that other potential URL codings are applicable, which however are beyond the scope of this study, since URL coding is the same for all classifiers (except for RNN), and the aim is to evaluate the role of classifiers in this learned data structure.
    \item \textit{Hardware}. The experiments were run on a Intel Core i5-8250U 1.60GHz Ubuntu machine with 16GB RAM. 
\end{itemize}



\subsection{Evaluation Framework}
\subsubsection{Classifiers screening}
Intuition suggests that the better a  classifier is at discriminating benign from malicious URLs, the better the performance of the Learned Bloom Filter using it. Nevertheless, the space occupied by the classifier plays a central role, and the trade-off performance/space is to be taken into account.

In order to shed quantitative  light on this aspect, we perform classification experiments involving the URLs and the classifiers, without the filter. 

\begin{itemize}
    \item \textit{Classifier validation}. Classifiers' generalization capabilities have been evaluated through a $5$-fold cross validation on the whole dataset, and the performance assessed in terms of Accuracy, $F1$ measures, and  space occupied. The latter includes the space of both classifier structure and input encoding, stored as a standard in Python via the serialization method {\tt dump} of library \textit{Pickle}~\cite{Pickle}. The results averaged across folds are calculated. In order to have a binary prediction necessary to compute Accuracy and $F1$, in this experiment the best threshold $\tau$ is estimated on the training set to maximize the $F1$.  
    \item \textit{Classifier model selection}. The best performing hyperparameter configuration  for each classifier (see Section~\ref{sub:classif}) is selected through a $5$-fold (inner) cross validation on the training set. In order to directly assess the trade-off Accuracy/Space, the key hyperparameters of the RNN and FFNN classifiers have been preset to $k=\{4, 8, 16\}$ (as done in~\cite{ma2020}), and $h=\{8, 16, 64\}$.
\end{itemize}
\subsubsection{Learned Bloom Filters}
Following~\cite{ma2020}, both typologies of learned Bloom Filters have been validated in the following holdout setting. The training regards the construction of the Bloom Filter on the malicious URLs, which follows a standard approach, but it must also account for the training of the classifier embedded in the filter.  Such a task is performed using  half of the   benign URL uniformly selected (in addition to the malicious ones), whereas the rest of benign URLs constitutes the holdout set used to compute the filter performance.  
\paragraph{Evaluation.} The filters have been also evaluated in terms of space, including both classifiers and auxiliary Bloom Filters space, and reject time. To this end, as baseline, the space and reject time of the classic Bloom Filter is also reported. As suggested in~\cite{Mitz18}, we tuned $\tau$ to achieve classifier false positive rates $\epsilon_{\tau}$ that are within the bounds required, and the performance of the filter evaluated for different admissible values of $\epsilon_{\tau}$ (see Section~\ref{sub:BFs}).

\begin{table}[t]
    \centering
\scriptsize
\begin{tabular}{|l|c|c|r|}
    \hline
    {\footnotesize     \textbf{Classifier} }&  {\footnotesize\textbf{Accuracy}} & {\footnotesize\textbf{F1} }& {\footnotesize\textbf{Space (Kb)} }\\
    \hline
    
    NB & $0.909 \pm 0.009$    & $0.522 \pm 0.003$  & $3.40$ \\
    SVM &  $0.929  \pm 0.001$  & $0.673 \pm 0.003$  &  $1.56$\\
    LR &  $0.930 \pm 0.001$  & $0.672  \pm  0.003$ & $ 1.61$\\
    RNN-16 & $0.935 \pm 0.003$ & $0.751 \pm 0.010$ & $ 9.45$ \\
    RNN-8 & $0.925 \pm 0.002$ & $0.733 \pm 0.005$ & $ 6.39$  \\
    RNN-4 & $0.916 \pm 0.009$ & $0.677 \pm 0.035$ & $ 5.51$  \\
    FFNN-64 & $0.962 \pm 0.001$ & $0.846 \pm 0.002$ & $ 21.80$ \\
    FFNN-16 & $0.952 \pm 0.001$ & $0.812 \pm 0.002$ & $ 5.63$ \\
    FFNN-8 & $0.946 \pm 0.001$ & $0.789 \pm 0.003$ & $ 2.94$  \\
    \hline
\end{tabular}	
\caption{Cross validation performance of the compared classifiers. For each measure also the standard deviation across folds is shown.}
\label{tab:classPerf}
\end{table}
\normalsize
 \begin{table}[t]
   \scriptsize
    \centering
    \begin{tabular}{|l|c|c|c|c|}
    \hline    
    {\small \textbf{$\epsilon$}} & 0.001 & 0.005 & 0.01 & 0.02\\[3pt]
    {\small \textbf{Space}} & $76.8$ & $58.9$ & $51.2$ & $43.5$ \\[3pt]
    {\small \textbf{Time}} & $1.22\cdot 10^{-6}$ & $1.03 \cdot 10^{-6}$ & $1.02 \cdot 10^{-6}$ & $1.00 \cdot 10^{-6}$ \\
    \hline
    \end{tabular}
    \caption{Size (in Kb) and average reject time (in sec.) of Bloom Filters storing malicious URLs by varying the false positive rate $\epsilon$.}\label{tab:BFsize}
\end{table}
\normalsize
\subsection{Results}
\subsubsection{Results of classifiers screening}
In Table~\ref{tab:classPerf} we report performance comparison of the adopted classifiers when trained on the whole dataset. 
FFNNs and RNNs achieve Accuracy and $F1$ values higher that the other competitors, with the former being preferable w.r.t. RNNs from all the standpoints (Accuracy, $F1$ and space/performance trade-off). These two families of classifiers are able to unveil (unlike  NB, SVM and LR) even non linear relationships between input and output, accordingly their superior  results are not surprising. The remaining classifiers are competitive in terms of Accuracy, less in terms of $F1$. Their best feature here is the compactness, being SVM and LR around $1.88\times$ and $13.97\times$ smaller than the smallest ($FFNN$-$8$) and biggest ($FFNN$-$64$)  NN-based models, respectively. 
This is a central issue in this setting, because the classifier size contributes to the total space of the learned Bloom Filter using it.  
To help this analysis, in Table~\ref{tab:BFsize} the size of the classical Bloom Filter with different desired false positive rates $\epsilon$ is shown. A large classifier, like $FFNN$-$64$, is practically not usable on this dataset, since it occupies, for example when $\epsilon = 0.02$, till half of the space of the filter. For this reason, in the next experiment, evaluating the performance of learned Bloom Filters, the range of values tested for $h$ is set to $\{10, 15, 20, 25, 30\}$.

\begin{table}[t]
\centering
\scriptsize
\begin{tabular}{|lc|c|c|c|}
    \hline
      & \multicolumn{4}{c|}{ {\small \textbf{$\epsilon$} }}\\
      Classifier &     0.001 &     0.005 &     0.010 &     0.020 \\
    \hline
    & \multicolumn{4}{c|}{LBF} \\
    \hline 
    NB &  3.12e-6 &  3.24e-6 &  3.01e-6 &  3.37e-6 \\
    SVM &  3.35e-6 &  3.23e-6&  3.43e-6 &  3.58e-6 \\
    LR &  3.77e-6 &  3.84e-6 &  3.95e-6 &  3.98e-6 \\
    RNN &  1.64e-3 &  1.75e-3 &  1.76e-3&  1.77e-3\\
    FFNN &  2.27e-4 &  2.73e-4 &  2.72e-4 & 2.73e-4\\
    \hline
    & \multicolumn{4}{c|}{SLBF}   \\
    \hline
    Bayes & 4.83e-6 &  4.90e-3 &  4.71e-6 &  5.03e-6 \\
    SVM &  4.32e-6 &  4.58e-6&  4.41e-6 &  4.59e-6 \\
    LR &  4.47e-6 &  4.64e-6&  4.35e-6 &  4.68e-6 \\
    RNN &  1.57e-3 &  1.71e-3 &  1.71e-3&  1.72e-3 \\
    FFNN &  2.70e-4 &  2.64e-4 &  2.63e-4 & 2.65e-4\\
    \hline    
\end{tabular}
\caption{Average reject time in seconds for one query of the different learned Bloom Filters.}\label{tab:time}
\end{table}
\subsubsection{Results of Bloom Filters comparison}
\begin{figure*}[t]
    \centering
    \includegraphics[width=0.99\textwidth]{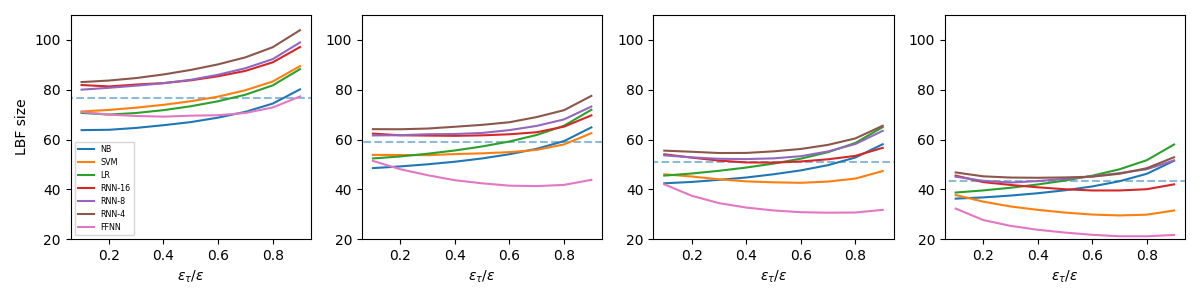}\\
    \includegraphics[width=\textwidth]{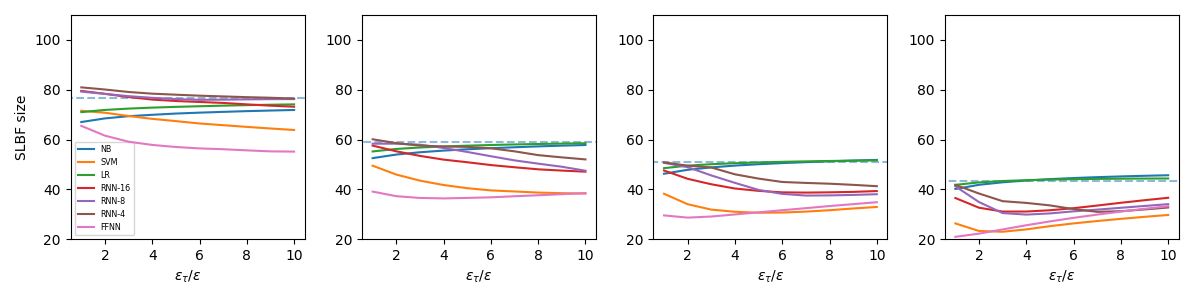}\\
    \caption{Size of LBF (first row) and SLBF (second row) for different classifier false positive rate. Dashed line is the size of the corresponding Bloom Filter. From the left, figures refer to a desired false positive rate $\epsilon$ of 0.001, 0.005, 0.01, 0.02.}
    \label{fig:LFB_SLBF}
\end{figure*}

Figure~\ref{fig:LFB_SLBF} depicts overall filter size when employing different classifiers and when varying both $\epsilon$ and $\frac{\epsilon_{\tau}}{\epsilon}$. Having the classifiers FFNN and RNN different choices for the performance/space trade-off, depending on the configuration of their key hyperparameters, here the one leading to the smallest learned filter has been reported. Specifically, $k=16$ and $h=20$. 

A first relevant emerging trend is that the choice of $\epsilon_{\tau}$ is critical to determine the best gain in space. In the case of LBFs, for small $\epsilon$ values  (first two columns), the best compression corresponds to low values of $\frac{\epsilon_{\tau}}{\epsilon}$, suggesting that the classifier precision (lower number of false positives) is more relevant when we aim at constructing LBFs with small false positive rates. Indeed, the filter size grows with $\frac{\epsilon_{\tau}}{\epsilon}$. On the other side, this means that when smaller $\epsilon$ values are required, the best choice is to rely more on the backup filter to guarantee a false positive rate $\epsilon$ (a classifier having a low $\epsilon_{\tau}$ produces more false negatives). Only the FFNN-based LBF with $\epsilon = 0.005$ already shows a trend observed also for larger $\epsilon$ values (last two columns), where for most LBFs the overall space decreases till a certain value of $\frac{\epsilon_{\tau}}{\epsilon}$ is attained, then it starts to increase. This means that the classifier plays a more crucial role in this cases, as confirmed by the higher space reduction with respect to the classic Bloom Filter (dashed line).

SLBFs work differently, and it is not intuitive to understand the reason of such a difference. We attempt to provide here some potential clues in this direction. A first and main difference, except for filters relying on NB and LR, for which the best $\epsilon_{\tau}$ is always the lowest one (probably due to the poor performance of NB and LR), is that now for smaller $\epsilon$ values the classifiers are allowed to produce more false positives, and conversely for larger choices of $\epsilon$.  Exactly as opposite to what happens for the LBFs. A possible explanation resides in the role of the initial Bloom Filter of the SLBF. Filtering out a portion of negatives allows the classifier to produce less false positives, thus fostering a different tuning of $\tau$. A second relevant difference with LBFs is the higher stability in terms of space with respect to the choice of $\tau$, which confirms the analyses in~\cite{Mitz18}. Overall, SLBFs tend to achieve larger size reduction than LBFs, mainly for smaller values of $\epsilon$, as a confirm of previous theoretical~\cite{Mitz18} and empirical~\cite{Dai} results.

Regarding the impact of individual classifiers on the overall size, in most cases the filter using a FFNN compresses the most with respect to the classic Bloom Filter, in accordance with the results shown in Table~\ref{tab:classPerf}; however, the SVM is very competitive, mainly when embedded in the SLBF, being almost indistinguishable with FFNN in terms of space gain. Another surprising result is for the LBFs with $\epsilon = 0.001$, where the NB allows the best space reduction. Such results are even more meaningful if we consider the reject times shown in Table~\ref{tab:time}: the filters built using SVMs and NB are two orders of magnitude faster than those using FFNNs.  
Thus, what is counterintuitive is that
non-linear or complex is not necessarily better than linear and simple when we are declining the characteristics of a classifier to be employed in a learned Bloom Filter. It is likely such results are tied to the size of key set, being higher the potential gain of a learned version vs.\ its classic counterpart when the number of keys increases~\cite{Dai,ma2020}. This is left to future investigations. 
%

\section{\uppercase{Conclusions and Future Developments}}
\label{sec:conclusion}
We have preliminarily investigated the impact of general purpose classifiers within recently proposed extension of Bloom Filters, named learned Bloom Filters. Following~\cite{Mitz18}, which emphasized how these learned extensions are dependent on the data distribution,  we have conducted an empirical evaluation in the context of malicious URLs detection, considering a wide range of general purpose classifiers.  Previous works on learned Bloom Filters have not focused on the appropriate choice of the classifier, and simply relied on non-linear models. Our results have confirmed the suitability of such classifiers, but also have affirmed that simpler classifiers (e.g., linear) might be the best choice in some specific contexts, especially when the reject time is also a key factor. 
Future works in this directions would extend and further validate the learned Bloom Filters on different application domains, e.g. Genomics, Cyber security, Web networking, and test the obtained results against problems with different key set sizes and distributions, thus allowing to unveil further insights into the role played by the classifiers, possibly including even those  discarded in this paper for their excessive space occupancy. Finally, the adoption of succinct representations of the trained classifiers, e.g. Neural Networks~\cite{NNcompress,SHAM,RRPR21}, might sensibly contribute to reduce the size of a learned filter.  
\vfill
\section*{\uppercase{Acknowledgements}}

\xblackout{This work has been supported by the Italian MUR PRIN project ``Multicriteria data structures and algorithms: from compressed to learned
indexes, and beyond'' (Prot. 2017WR7SHH). Additional support to R.G. has been granted by Project INdAM - GNCS  ``Analysis and Processing of Big Data based on Graph Models''.} 

\bibliographystyle{apalike}
{\small
\bibliography{main}

\begin{thebibliography}{}

\bibitem[Amato et~al., 2021]{amato2021learned}
Amato, D., Giancarlo, R., and Bosco, G.~L. (2021).
\newblock Learned sorted table search and static indexes in small space:
  Methodological and practical insights via an experimental study.
\newblock {\em CoRR}, abs/2107.09480.

\bibitem[Bloom, 1970]{Bloom70}
Bloom, B.~H. (1970).
\newblock Space/time trade-offs in hash coding with allowable errors.
\newblock {\em Commun. ACM}, 13(7):422--426.

\bibitem[Boffa et~al., 2021]{Boffa21}
Boffa, A., Ferragina, P., and Vinciguerra, G. (2021).
\newblock A ``learned'' approach to quicken and compress rank/select
  dictionaries.
\newblock In {\em Proceedings of the SIAM Symposium on Algorithm Engineering
  and Experiments (ALENEX)}.

\bibitem[BOTW, 2021]{URLBotw}
BOTW (2021).
\newblock {Best of the Web -- Free Business Listing}.
\newblock \url{https://botw.org}.
\newblock Last checked on Oct.~18, 2021.

\bibitem[Broder and Mitzenmacher, 2002]{Broder2005}
Broder, A. and Mitzenmacher, M. (2002).
\newblock {Network Applications of Bloom Filters: A Survey}.
\newblock In {\em Internet Mathematics}, volume~1, pages 636--646.

\bibitem[Cho et~al., 2014]{RNN}
Cho, K., van Merrienboer, B., G{\"{u}}l{\c{c}}ehre, {\c{C}}., Bahdanau, D.,
  Bougares, F., Schwenk, H., and Bengio, Y. (2014).
\newblock Learning phrase representations using {RNN} encoder-decoder for
  statistical machine translation.
\newblock In {\em Proc.\ of the 2014 Conf.\ on Empirical Methods in Natural
  Language Processing, {EMNLP} 2014, October 25-29, 2014, Doha, Qatar, {A}
  meeting of SIGDAT, a Special Interest Group of the {ACL}}, pages 1724--1734.
  {ACL}.

\bibitem[Cortes and Vapnik, 1995]{SVM}
Cortes, C. and Vapnik, V. (1995).
\newblock Support-vector networks.
\newblock {\em Machine learning}, 20(3):273--297.

\bibitem[Cox, 1958]{LRegr}
Cox, D.~R. (1958).
\newblock The regression analysis of binary sequences.
\newblock {\em Journal of the Royal Statistical Society: Series B
  (Methodological)}, 20(2):215--232.

\bibitem[Dai and Shrivastava, 2020]{Dai}
Dai, Z. and Shrivastava, A. (2020).
\newblock {A}daptive {L}earned {B}loom {F}ilter ({A}da-{BF}): Efficient
  utilization of the classifier with application to real-time information
  filtering on the web.
\newblock In {\em Advances in Neural Information Processing Systems},
  volume~33, pages 11700--11710. Curran Associates, Inc.

\bibitem[Duda and Hart, 1973]{NaiveBC}
Duda, R.~O. and Hart, P.~E. (1973).
\newblock {\em Pattern Classification and Scene Analysis}.
\newblock John Willey \& Sons, New Yotk.

\bibitem[Duda et~al., 2000]{duda20}
Duda, R.~O., Hart, P.~E., and Stork, D.~G. (2000).
\newblock {\em Pattern Classification, 2nd Edition}.
\newblock Wiley.

\bibitem[Ferragina et~al., 2021]{FERRAGINA21}
Ferragina, P., Lillo, F., and Vinciguerra, G. (2021).
\newblock On the performance of learned data structures.
\newblock {\em Theoretical Computer Science}, 871:107--120.

\bibitem[Ferragina and Vinciguerra, 2020a]{Ferragina:2020book}
Ferragina, P. and Vinciguerra, G. (2020a).
\newblock Learned {D}ata {S}tructures.
\newblock In {\em Recent {T}rends in {L}earning {F}rom {D}ata}, pages 5--41.
  Springer International Publishing.

\bibitem[Ferragina and Vinciguerra, 2020b]{Ferragina:2020pgm}
Ferragina, P. and Vinciguerra, G. (2020b).
\newblock The {PGM-index}: a fully-dynamic compressed learned index with
  provable worst-case bounds.
\newblock {\em {PVLDB}}, 13(8):1162--1175.

\bibitem[Freedman, 2005]{FreedmanStat}
Freedman, D. (2005).
\newblock {\em Statistical {M}odels : {T}heory and {P}ractice}.
\newblock {Cambridge University Press}.

\bibitem[Kipf et~al., 2020]{Kipf20}
Kipf, A., Marcus, R., van Renen, A., Stoian, M., Kemper, A., Kraska, T., and
  Neumann, T. (2020).
\newblock Radixspline: A single-pass learned index.
\newblock In {\em Proc.\ of the Third International Workshop on Exploiting
  Artificial Intelligence Techniques for Data Management}, aiDM '20, pages
  1--5. Association for Computing Machinery.

\bibitem[Kraska et~al., 2018]{Kraska18}
Kraska, T., Beutel, A., Chi, E.~H., Dean, J., and Polyzotis, N. (2018).
\newblock The case for learned index structures.
\newblock In {\em Proc.\ of the 2018 Int.\ Conf.\ on Management of Data},
  SIGMOD '18, pages 489--504, New York, NY, USA. Association for Computing
  Machinery.

\bibitem[Long et~al., 2019]{NNcompress}
Long, X., Ben, Z., and Liu, Y. (2019).
\newblock A survey of related research on compression and acceleration of deep
  neural networks.
\newblock {\em Journal of Physics: Conference Series}, 1213:052003.

\bibitem[Ma and Liang, 2020]{ma2020}
Ma, J. and Liang, C. (2020).
\newblock An empirical analysis of the learned bloom filter and its extensions.
\newblock Unpublished. Paper and code no more available on line.

\bibitem[{Machine Learning Lab}, 2021]{URLTrento}
{Machine Learning Lab} (2021).
\newblock Hidden fraudulent urls dataset.
\newblock
  {https://machinelearning.inginf.units.it/data-and-tools/hidden-fraudulent-urls-dataset}.
\newblock Last checked on Oct.~18, 2021.

\bibitem[Maltry and Dittrich, 2021]{Mailtry21}
Maltry, M. and Dittrich, J. (2021).
\newblock A critical analysis of recursive model indexes.
\newblock {\em CoRR}, abs/2106.16166.

\bibitem[Marcus et~al., 2020a]{Marcus20}
Marcus, R., Kipf, A., van Renen, A., Stoian, M., Misra, S., Kemper, A.,
  Neumann, T., and Kraska, T. (2020a).
\newblock Benchmarking learned indexes.
\newblock {\em arXiv preprint arXiv:2006.12804}, 14:1--13.

\bibitem[Marcus et~al., 2020b]{Markus20b}
Marcus, R., Zhang, E., and Kraska, T. (2020b).
\newblock {CDFS}hop: Exploring and optimizing learned index structures.
\newblock In {\em Proc.\ of the 2020 ACM SIGMOD Int.\ Conf.\ on Management of
  Data}, SIGMOD '20, pages 2789--2792.

\bibitem[Marin{\`o} et~al., 2021a]{SHAM}
Marin{\`o}, G.~C., Ghidoli, G., Frasca, M., and Malchiodi, D. (2021a).
\newblock Compression strategies and space-conscious representations for deep
  neural networks.
\newblock In {\em Proceedings of the 25th International Conference on Pattern
  Recognition (ICPR)}, pages 9835--9842.
\newblock doi:10.1109/ICPR48806.2021.9412209.

\bibitem[Marin{\`o} et~al., 2021b]{RRPR21}
Marin{\`o}, G.~C., Ghidoli, G., Frasca, M., and Malchiodi, D. (2021b).
\newblock Reproducing the sparse huffman address map compression for deep
  neural networks.
\newblock In {\em Reproducible Research in Pattern Recognition}, pages
  161--166, Cham. Springer International Publishing.
\newblock doi:10.1007/978-3-030-76423-4\_12.

\bibitem[Mitzenmacher, 2018]{Mitz18}
Mitzenmacher, M. (2018).
\newblock A model for learned bloom filters and optimizing by sandwiching.
\newblock In {\em Advances in Neural Information Processing Systems},
  volume~31. Curran Associates, Inc.

\bibitem[Mitzenmacher and Vassilvitskii, 2020]{Mitz20}
Mitzenmacher, M. and Vassilvitskii, S. (2020).
\newblock Algorithms with predictions.
\newblock {\em CoRR}, abs/2006.09123.

\bibitem[{Python Software Foundation}, 2021]{Pickle}
{Python Software Foundation} (2021).
\newblock pickle -- python object serialization.
\newblock \url{https://docs.python.org/3/library/pickle.html}.
\newblock Last checked on Oct.~18, 2021.

\bibitem[UNIMAS, 2021]{URLUnimas}
UNIMAS (2021).
\newblock Phishing dataset.
\newblock \url{https://www.fcsit.unimas.my/phishing-dataset}.
\newblock Last checked on Oct.~18, 2021.

\bibitem[Vaidya et~al., 2021]{Kraskap}
Vaidya, K., Knorr, E., Kraska, T., and Mitzenmacher, M. (2021).
\newblock Partitioned learned bloom filters.
\newblock In {\em International Conference on Learning Representations}.

\bibitem[Zell, 1994]{FFNN}
Zell, A. (1994).
\newblock {\em Simulation neuronaler Netze}.
\newblock habilitation, Uni Stuttgart.

\end{thebibliography}
}

\end{document}